\documentclass{article}
\usepackage{aaai19}
\usepackage[utf8]{inputenc}
\usepackage{layout}

\usepackage{times}  
\usepackage{helvet} 
\usepackage{courier}  
\usepackage[hyphens]{url}  
\usepackage{hyperref}
\usepackage{natbib}
\usepackage[dvipsnames]{xcolor}
\usepackage{graphicx}
\usepackage{url}            
\usepackage{booktabs}       
\usepackage{amsfonts}       
\usepackage{nicefrac}       
\usepackage{microtype}      

\usepackage{latexsym}
\usepackage{amsmath}
\usepackage{amsfonts}
\usepackage{mathtools}
\usepackage{url}
\usepackage{graphicx}
\usepackage{appendix}
\usepackage{multirow}
\usepackage{color}
\usepackage{comment}
\usepackage[most]{tcolorbox}
\usepackage{makecell}

\definecolor{ourdarkblue}{HTML}{092E6B}

\tcbset{nobeforeafter, bottom=1mm, top=1mm, left=1mm, right=1mm, frame hidden, boxrule=0pt, arc=3mm, colframe=white}

\newcolumntype{C}{>{\centering\arraybackslash}p{0.3\textwidth}}

\title{Open-Domain Conversational Agents: \\Current Progress, Open Problems, and Future Directions}
\author{Stephen Roller\thanks{Equal contribution to this position paper.}, Y-Lan Boureau$^{*}$, Jason Weston$^{*}$, Antoine Bordes, Emily Dinan, \\
\textbf{\Large Angela Fan, David Gunning, Da Ju, Margaret Li, Spencer Poff, Pratik Ringshia,}\\
\textbf{\Large Kurt Shuster, Eric Michael Smith, Arthur Szlam, Jack Urbanek, Mary Williamson}\\
Facebook AI Research\\
\tt{parlai@fb.com}}

\begin{document}

\maketitle

\begin{abstract}
We present our view of what is necessary to build an engaging
open-domain conversational agent: covering  the qualities of such an agent,
the pieces of the puzzle that have been built so far,
and the gaping holes we have not filled yet. We present a biased view, 
focusing on work done by our own group,
while citing related work in each area.
In particular, we discuss in detail the 
properties of continual learning,
providing engaging content, and being well-behaved -- and how to 
measure success in providing them.
We end with a discussion of
our experience and learnings, and our recommendations to the community.
\end{abstract}

Good open-domain conversationalists seamlessly blend entertaining wit and knowledge while making others feel heard. The breadth of possible conversation topics and lack of a well-defined objective make it challenging to define a roadmap towards training a good conversational agent, or chatbot. Despite recent progress across the board \citep{adiwardana2020meena,roller2020recipes}, conversational agents are still incapable of carrying an open-domain conversation that remains interesting, consistent, accurate, and reliably well-behaved (e.g., not offensive) while navigating a variety of topics.

Traditional task-oriented dialogue systems rely on slot-filling and structured modules (e.g., \citet{young2013pomdp,gao2019neural,jurafsky2019speech}). These approaches have proven
adept at producing usable commercial systems in narrow domains such as plane ticket booking. However, they are limited to the domain they were trained on and do not afford generalization to new domains or open chit-chat settings, necessitating the coding of many modules, or skills, and a managing system that switches between them. End-to-end approaches based
on neural networks, on the other hand, offer the promise of adapting to arbitrarily wide new domains without additional handcrafting, but have yet to reach the full potential promised. 
Deep architectures trained end-to-end have been very successful in many other 
domains, such as speech
recognition \citep{hinton2012deep,collobert2016wav2letter}, computer vision \citep{krizhevsky2012imagenet}, 
and machine translation \citep{sutskever2014sequence,gehring2017convolutional}. 
Hence, the research community is investing heavily in improving end-to-end models for dialogue \citep{zhang2019dialogpt,adiwardana2020meena,roller2020recipes},
in the hope of achieving  similar success. 

In this paper, we highlight some of the recent work towards that goal, 
beginning by attempting  to define the problem itself.
We thus describe the desirable traits that we believe a superhuman open-domain conversational agent should have, state principles our research follows, 
and propose ways to measure our progress. We examine the challenges of this research program, summarize the results we have already obtained, and propose guidelines for the community to accelerate progress.
Note that while we try to cite related work where possible, this article is written with
 a strong bias toward describing the progress in the goals 
and research directions of our own group.
Further,  we discuss only open academic research with reproducible published results, 
hence we will not address much of the considerable work  
that has been put into building commercial systems, where methods, data and results are not in the public domain.
Finally, given that we focus on open-domain conversation, we do not focus on specific goal-oriented techniques; we also do not cover spoken dialogue in this work, focusing on text and image input/output only.
For more general recent surveys, see \citet{gao2019neural,jurafsky2019speech,huang2020challenges}.

\section{Qualities of a Conversational Agent}

We define our long-term goal as building a superhuman open-domain conversational agent. 
That is, we aim to build an agent that is preferred on average to an alternative human speaking partner in open conversation, which we will discuss in detail later in the {\em measuring success} section
(evaluation being an open problem in itself). We note that this is different  
from passing a Turing test \citep{turing1950computing}: we do not wish to fool humans into believing our agent is human, but instead that our agent should be enjoyed as a speaking partner, which we believe is a simpler problem.

We expect that such an agent must be \emph{continually-learning}, must provide \emph{engaging content} during conversations, and should be \emph{well-behaved}. Each of these high-level traits can be unpacked into a number of behaviors and attributes, many of which constitute major research directions with open questions. We describe each of the main properties in turn, along with their major challenges.

\subsection{Continually Learning}

Continual learning is a cornerstone of the conversational agent we envision, allowing it to adapt to new contexts,
new users, keep up to date with current conversation
topics, and continuously improve.
This entails three primary skills:
continuous online training of underlying models, extracting
useful learning signals from interaction,
and updating relevant sources of knowledge.

\paragraph{Continual Online Training}
Recent dialogue research has leveraged
various data sources for training: corpora of human-human conversations in narrow domains (see \citet{serban2018survey} and the list of currently available ParlAI tasks\footnote{\url{https://parl.ai/docs/tasks.html}} for a large set of available corpora), public conversations on internet discussion boards or social media, 
or acted dialogues from crowdsourced workers \citep{zhang2018personalizing,dinan2018wizard,shuster2018engagingimagechat,rashkin2019empathetic,shuster2019dialogue,smith2020bst}.
Relying on static datasets allows for reproducibility and model comparisons, 
but creates a potential mismatch of
data distribution between train time and deployment of the conversational agent.
The framework of combining a pre-trained
model with fine-tuning over another
dataset of interest has generally produced good
results, as we have seen in much of our work \citep{dinan2019safety,dinan2019second,dinan2018wizard,humeau2019polyencoder,rashkin2019empathetic,zhang2019dialogpt,shuster2019dialogue,smith2020bst,roller2020recipes}. However, this fine-tuning
procedure over a static dataset still does not
allow for dynamically adapting to new topics of
interest or new audiences, and the current set
of available tasks is far from covering
everything that an open-domain conversation
might touch on.
Thus, an important part of our 
research program consists in deploying
conversational agents so that 
new data of humans interacting with
the agent in an open-ended conversation
can continuously be obtained and used
for fine-tuning models with fresh data.

{\em Open problems}. Obtaining a source of
continually renewed data opens the door to many open problems. The general challenges of
never-ending learning \citep{mitchell2018never,carlson2010toward} and avoiding catastrophic forgetting \citep{french1999catastrophic,kirkpatrick2017overcoming} have a particular flavor in the
domain of conversation: the general task is always to talk to people about any subject, but the set of people, the topics of interest, the facts that have happened may all change. As in other domains, the performance that can be achieved
on a task after fine-tuning is highly dependent on the
data a model has been pre-trained on. Empirical
evidence suggests that data with a very large range of 
domains (e.g., social media data) provides the best basis for fine-tuning \citep{zhang2019dialogpt,shuster2019dialogue,adiwardana2020meena,roller2020recipes}.
It remains unclear how to elicit interaction data that
would be the most useful as a general-purpose pre-training
corpus. Comparisons across many different types
of pre-training suggest that training on many
existing dialogue corpora is not as effective
across the board as simply using readily available
non-interactive corpora \citep{shuster2019dialogue}. The same may be 
true of any interactive data we collect in
the context of a given framework.
Continual training also requires figuring out the best
trade-off between being able to stay current (biasing towards
more recent data) and retaining ability to talk about past topics. There could be many different policies around this,
and reinforcement learning could be used to explore and
optimize to find the best successful ones, in terms of
people's satisfaction. More generally, a continually learning
agent is a good substrate for comparing policies targeting
all kinds of desirable objectives such as all the traits
mentioned in this overview. 
However, determining what a
good reinforcing signal should be for open-domain conversation
is very much  still an open problem: contrary to goal-oriented
dialogue, where there is a clear sense of what the reward is,
open-domain conversation does not have natural rewards.
Implicit signals such as dialogue length and frequency of
engagement have been used as metrics to rate models, for example for the Alexa prize~\citep{ram2017alexaprize}, but capture 
the ability to create a sticky habit rather
than the quality of the interaction per se.
Repeatedly asking users for satisfaction or detailed quality
ratings is cumbersome and decreases the fluidity of the interaction. Some attempts at predicting a quality score from a range of different signals have shown some positive results~\citep{serban2017deep,fang2018sounding,ghandeharioun2019approximating}, and have been used to train models through reinforcement learning~\citep{serban2017deep,fang2018sounding}, but they still 
show limited correlation with gold standard
quality ratings~\citep{serban2017deep}.
This approach leads to the next topic -- how to learn
from interaction directly in the conversation rather than
from a separate rating functionality.

 \begin{figure}
     \centering
     \includegraphics[width=\linewidth]{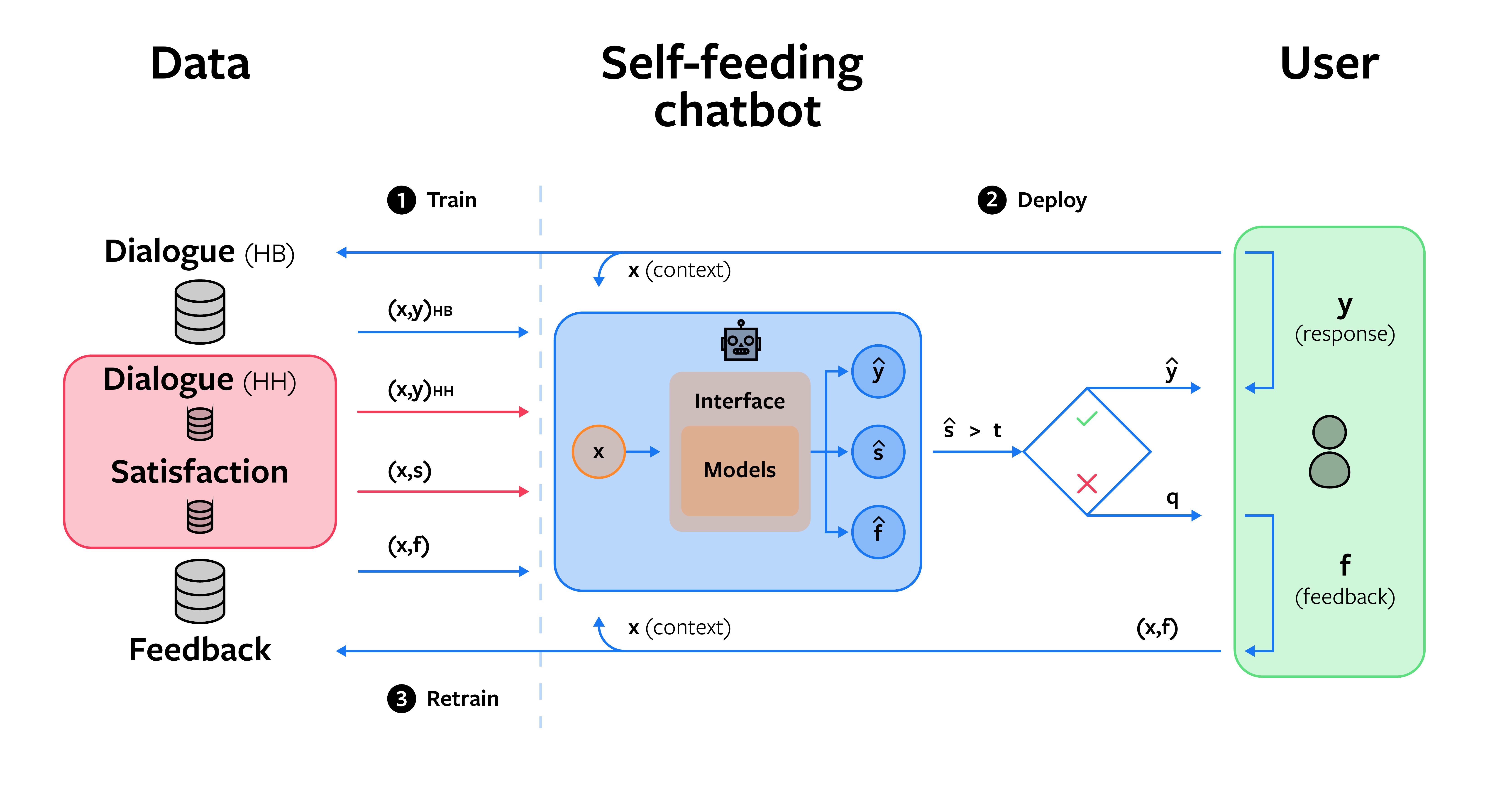}
     \caption{The self-feeding chatbot trains on the dialogues it engages in to continually learn \citep{hancock2019selffeeding}.}
     \label{fig:self_feeding}
 \end{figure}

\paragraph{Learning from interaction}
Beyond training on additional in-distribution data in a self-supervised way,
an exciting avenue for refining conversational agents consists in 
taking advantage of the interactive
character of a conversation by directly
soliciting feedback from conversation
partners. This has been explored in \citet{hancock2019selffeeding}, where
a ``self-feeding chatbot" learns to
estimate its partner's satisfaction
and can also ask for feedback when
it believes it has made a mistake, see Figure~\ref{fig:self_feeding}. Learning from these additional signals
significantly improves performance,
especially when the amount of data
is initially small. 
There has also been some work on predicting
conversation quality from signals such as sentiment or 
questions \citep{ghandeharioun2019approximating} or using other forms of feedback from the conversation \citep{li2016dialogue,li2016learning,weston2016dialog,yang2017mastering}, but
results are still preliminary.

{\em Open problems}. Feedback that is not explicitly requested by the conversational agent is not clearly marked as being about
the conversation itself and not about the subject of the conversation. For example, detecting a negative sentiment could mean that the conversation partner is upset about something
that happened to them outside of the conversation, and the
appropriate response would then be to display empathy, not figure
out how the conversation generation went wrong.
Another open problem is how to use sentiment or other signals
inferred from the conversation in a reinforcement learning
setting, as an objective of open-domain conversation.
An example would be to try and avoid offensive comments or elicit positive sentiment. 
The objective of obtaining a given emotional response from an artificial agent
was in fact used in a recent paper leveraging reinforcement 
learning in conversation grounded
in a fantasy adventure game \citep{prabhumoye2020}.
But there are many difficulties when it comes to optimizing
directly over interaction rewards, or proxy
automated metrics: models could try and
dissuade conversation partners from talking about anything
negative, or only focus on topics that are known to be more
positive, rather than being truly generalist.
Models optimizing a proxy metrics could simply lead
to those metrics becoming artificially inflated
and gradually decoupled from the true underlying 
 quality of the conversation.

\paragraph{Updating sources of knowledge}

A tip frequently given to people aiming
to become better conversationalists is
to consult the news to know what is currently happening. Conversation 
topics shift according to current
events or trends, and a good conversational
agent needs to be able to adapt to these
trends. This could be achieved through 
a variety of ways. If the agent has been trained to retrieve and incorporate information from an external
source of knowledge \citep{dinan2018wizard,qin2019conversing,prabhumoye2019towards,ghazvininejad2017knowledge}, then simply updating that source would
allow the agent to inject current information
into the conversation. If the source is itself dynamic
(e.g., Wikipedia is constantly being updated), then
simply reading from the updated version could be enough.

An important consideration when striving 
to stay current is that this may put
retrieval models at a disadvantage.
Pure retrieval models produce utterances
by retrieving from a set of training
utterances. This precludes saying anything
that was not said before the time when 
the set of retrieval utterances was created.
Generative models build new utterances
from scratch and are therefore not
subject to that limitation, and they
are naturally better suited to adapting
to changing contexts. 
Until recently, their performance was below that of
retrieval models  \citep{dinan2018wizard,li2019acute,rashkin2019empathetic,shuster2018imagecaption,zhang2018personalizing}, unless they relied on refining
retrieved utterances \citep{weston2018retrieve}.
However, larger pre-training datasets coupled with improved decoding choices, such as imposing a minimum length constraint on decoded generations, has been shown to erase the superiority of retrieval models \citep{roller2020recipes}, and generative models are now being rated highly by humans \citep{adiwardana2020meena,roller2020recipes}
.

{\em Open problems}. The very nature of the challenge of staying
current makes it difficult to devise a suitable benchmark, as a static benchmark does not capture the ability of adapting to
changing topics. Some measure of that can be achieved through partitioning data and topics between training and validation/test \citep{dinan2018wizard}, but this only works in settings where there is a clear delineation of topics, as opposed to the more fluid nature of natural chitchat, and importantly, does not address
the important point of deciding what topics are interesting to introduce in a conversation in the first place.
However there are already works in the space
of conversational AI that show promise for gauging an agent's
ability to adapt. \citet{dinan2019safety} suggests a protocol
for a dynamically evolving benchmark; the same idea could
be adapted to gauge what topics a conversation partner
expects to be able to discuss with an agent, and update
the agent accordingly.
The need for a dynamic benchmark is also a potential
advantage of deploying conversational agents for wide-spread
interaction with people: a dynamic benchmark could then be defined as a
regular survey of people who engage with a released conversational agent, for example asking them to rate whether
the agent was capable of conversing about the topics that 
they were interested in. 

\paragraph{Interacting with human users}

Our approach to achieving continual learning at scale relies on large-scale interaction with humans, which in turn requires our systems to be fielded and suitable for interaction with willing human users.
To that end, it is important to
make conversational systems resilient and capable of
handling many human conversation partners.
In particular, systems are easier to 
deploy and train in a continual manner if they are computationally 
reasonable. Architectures proposed in \citet{humeau2019polyencoder}
achieve promising trade-offs that
maintain high performance while 
allowing for substantial computational
savings.
As for connecting conversational agents
to human conversation partners, we have
deployed a conversational agent as a 
publicly available game\footnote{\url{https://parl.ai/projects/beat_the_bot}}
dual goal of collecting
human utterances as additional examples of good conversations, and obtaining continuous
human evaluation of new conversational
models. The game is structured as a
voting game where a human is paired with
another human and is asked to write response utterances, as well as select between
utterances from the model and utterances
written by the other human player.

{\em Open problems}. While a lot of progress has been made
in making language architectures more compact, the best-performing
systems for end-to-end open conversation are still relying
on memory- and compute-heavy Transformer architectures \citep{adiwardana2020meena,roller2020recipes}. 
Quantizing models and diminishing their memory and computation
footprint is an exciting problem space, and could ultimately
allow people to interact with an on-device model. Recent work on creating smaller architectures through knowledge distillation~\citep{sanh2019distilbert}, adaptive spans~\citep{sukhbaatar2019adaptive}, and pruning~\citep{fan2019reducing} provide promising directions for creating compact but high performance models.

\subsection{Engaging Content}

Humans will not want to interact with an agent unless it provides 
content that engages them in its  messages.
In a goal-oriented setting (e.g., a weather forecast) this is minimally supplied by achieving the goal, however  even
in those settings, and especially in others without such clear goals, 
there are multiple important factors that are at play. 
We cover some of those issues here.

 \begin{figure}
     \centering
     \includegraphics[width=\linewidth]{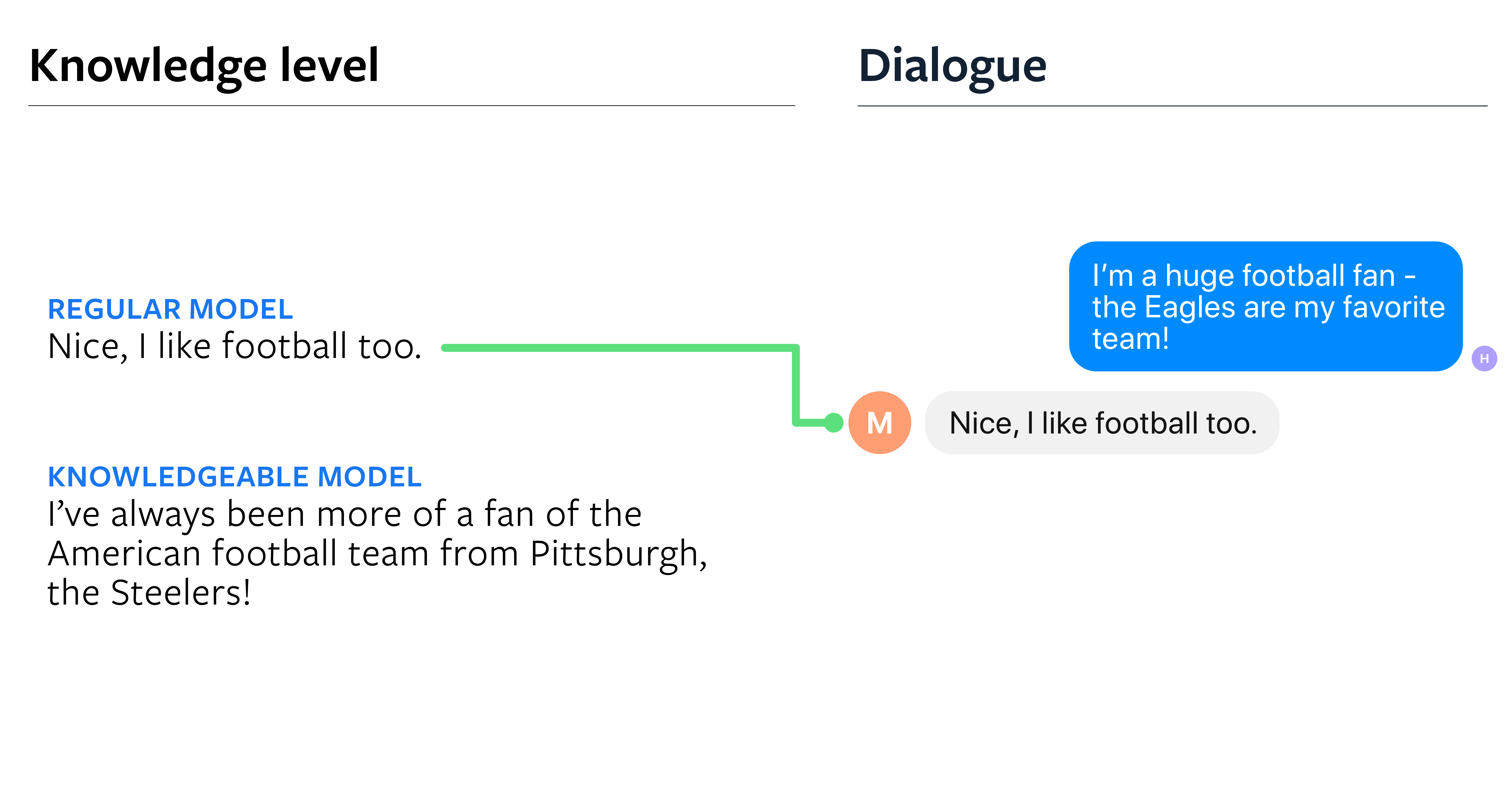}
     \caption{Using Knowledge: the Wizard of Wikipedia task \citep{dinan2018wizard}}.
 \label{fig:wiz-example}
 \end{figure}

\paragraph{Expert \& Knowledgeable}

Firstly, it is important that a general conversationalist exhibit a broad familiarity with different experiences or common background knowledge, or else as a specialist conversationalist have in-depth expertise in the skill demanded.
 In order to discuss with an art lover, an agent should command a reasonable level of knowledge about what are some famous pieces, techniques, or artists. A science geek would similarly require some information about space. An agent should exhibit the ability to work with knowledge and facts, and incorporate this information skillfully into its replies. 

Traditional goal-oriented dialogue has 
focused on narrow tasks that would typically 
be useful for a dialogue-based assistant, for example
restaurant \citep{henderson2014second}, taxi, train, and hotel \citep{budzianowski2018multiwoz}
or trip \citep{asri2017frames} booking. 
Classical goal-oriented dialogue literature typically uses structured knowledge,
slot filling or labeling, and studies reinforcement learning
extensively \citep{singh2000reinforcement}.

Question answering (QA) is another area where agents can display their expertise, 
typically recalling knowledge from large structured or unstructured resources
and then formulating a response \citep{chen2017reading,fan2019eli5}. 
Recent QA datasets have extended
to a conversational form with a series of questions possibly referencing
earlier ones in the conversation \citep{choi2018quac,reddy2019coqa}.

However, neither goal-oriented nor QA tasks completely  cover 
what a knowledgeable open-domain conversational agent should be able to do. 
To that end, human-human dialogues where people discuss topics in
depth have also been studied. In Wizard of Wikipedia  \citep{dinan2018wizard}
 22k such dialogues were collected between an expert partner and a curious learner (see also \citet{ghazvininejad2017knowledge,parthasarathi2018extending} for some other related datasets). 
To do this, 1k conversational topics were first crowdsourced, ranging from armadillos to ice cream to lifeguards, and then each dialogue starts with a chosen topic from among them. The expert speaker has access to a retrieval search engine over Wikipedia with the last dialogue turns as the query, and can thus inject 
knowledge obtained from there into the conversation. The aim of collecting the data in this way is one can then make this available to a conversational agent that learns to replace the human expert instead.
A new architecture, called Transformer Memory Networks, was designed which yields more knowledgeable agents, outperforming systems that do not employ a memory structure for storing knowledge in both automatic metrics and human evaluations. Generative model variants yield the most pronounced improvement and are rated by humans as 26\% more engaging on average than their knowledgeless counterparts.

Our eventual hope is to combine these skills -- open domain knowledgeable conversation, QA and task completion amongst others -- to build
a truly engaging, knowledgeable and skillful bot.
 
{\em Open problems}. Being expert and knowledgeable
is connected fundamentally to both {\em memory} and
{\em reasoning}, especially commonsense reasoning, which we discuss in the separate 
sections to follow. 
While we have made progress on individual problems, e.g. specific  tasks
or question answering in general,
we are still missing a strong ability to transfer to
new tasks, one of the most fundamental open problems in machine learning today. 
We believe this will be solved by improvements in both (i) architectures and learning mechanisms that better incorporate compositionality; and (ii) continual learning that updates  knowledge and expertise in those tasks.

\paragraph{Expressiveness and Flow}

Maintaining a good conversation requires balance -- between simplicity and detail; staying on topic and changing it; asking questions and answering them.
In generative models, a known issue is their 
propensity to produce short, dull utterances which over-use frequent words, and under-use
rare words -- which does not follow the human training distribution \citep{holtzman2019curious,fan2018hierarchical}. 
Meanwhile at the discourse, rather than single utterance level, the training procedures
typically employed are even less well suited -- as the classical next token prediction objective is far from planning an entire dialogue flow. Thus, approaches that simply optimize perplexity 
might fail to ask any questions of the user, for example, or can repeat the same topic over and over \citep{dinan2019second}.

\begin{figure}[t]
    \centering
     \includegraphics[width=\linewidth]{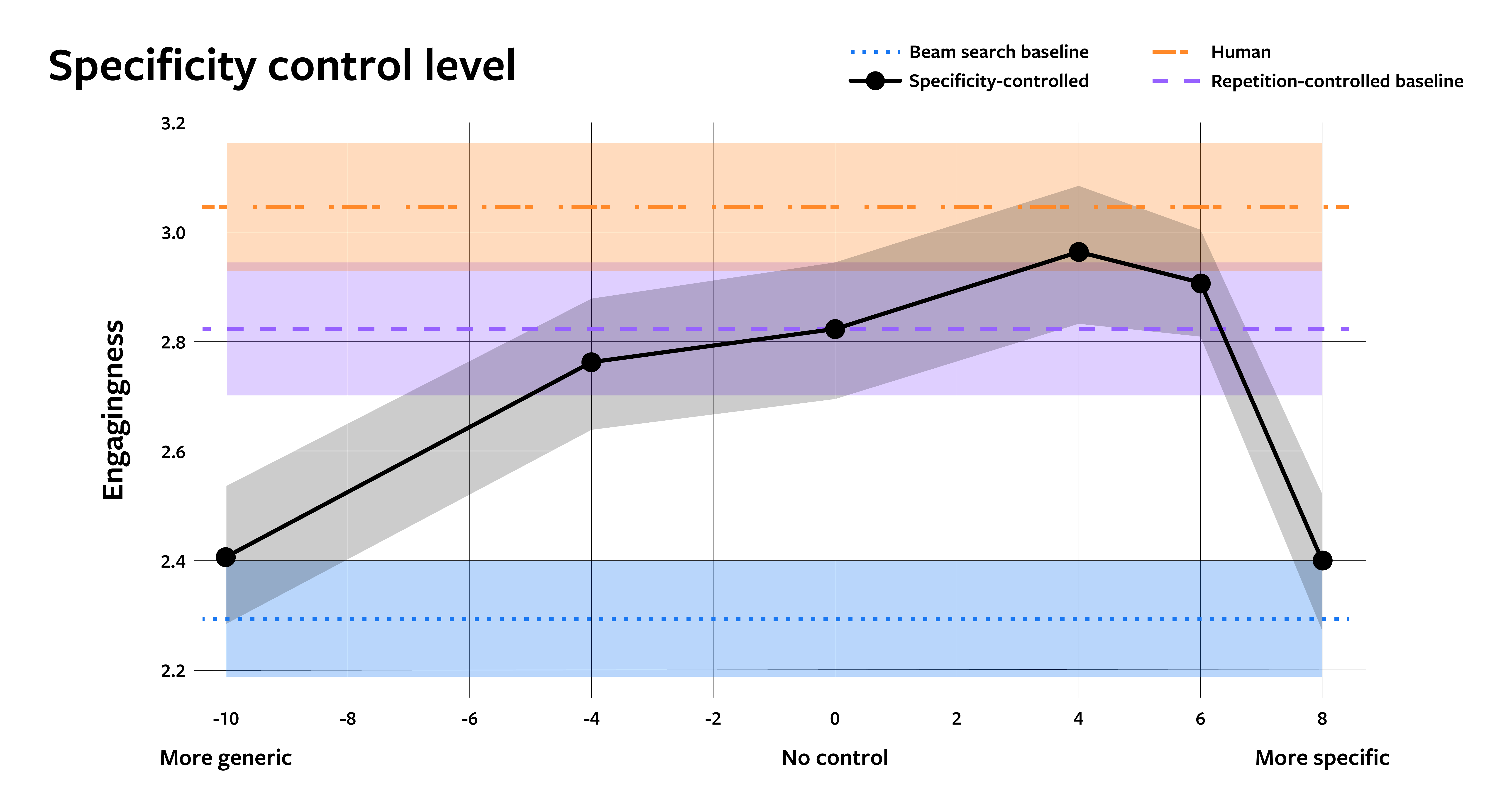}
     \caption{Controlling specificity in generative models affects user engagingness evaluations \citep{see2019goodconversation}.}
     \label{fig:specificity_graph}
\end{figure}

\begin{figure}
     \centering
     \includegraphics[width=\linewidth]{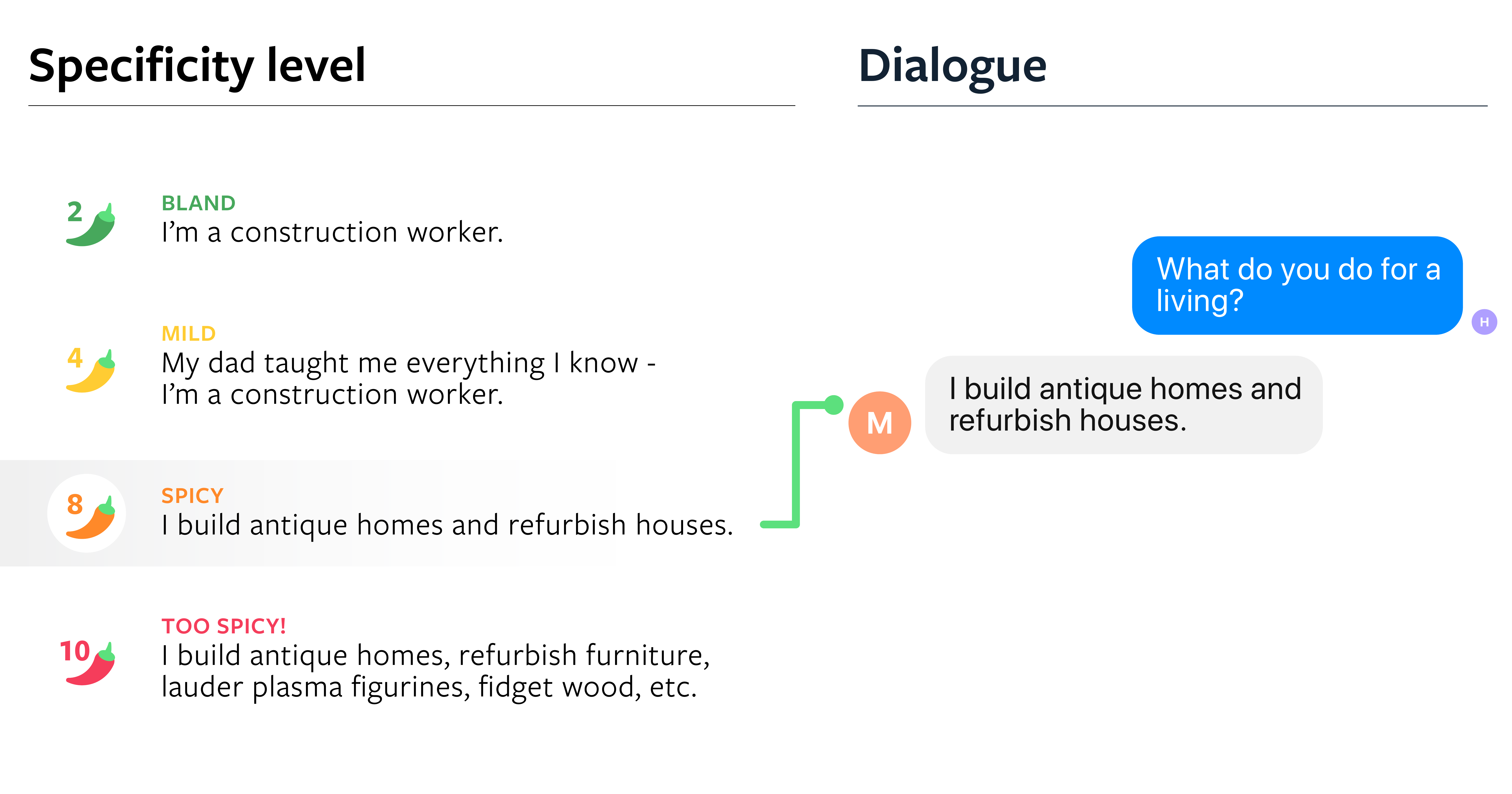}
     \caption{Specificity level using generative control \citep{see2019goodconversation}.}
     \label{fig:specificity}
\end{figure}

Numerous works have attempted to work on the so-called generic response problem. 
One solution is the use of 
controllable neural text generation methods, in particular conditional training 
\citep{fan2018controllable, kikuchi2016controlling, peng2018towards} and weighted decoding \citep{ghazvininejad2017hafez,baheti2018generating}. These methods provide a mechanism to control, and hence 
increase, the rare words used, resulting in less generic utterances.
In the work of \citet{see2019goodconversation}, it was
shown that controlling for such a measure strongly affects engagingness according to human evaluations, see Figures \ref{fig:specificity_graph} and \ref{fig:specificity}.

The work of \citet{see2019goodconversation} goes further and shows that
it is possible to control multiple important attributes for chitchat dialogue: repetition, specificity, response-relatedness and question-asking, in order
to optimize for well-balanced conversations. Human evaluations measured the effect of these control parameters on multi-turn interactive conversations on the PersonaChat task,
and showed repetition and question-asking were also similarly controllable, and importantly, provide clear improvements in human quality judgments. The final model is one of the best approaches on this task \citep{li2019acute}.

Another recent approach is the use of so-called unlikelihood training \citep{welleck2019neuraltext}. Intended as a replacement to classical likelihood training, 
it aims to fix the problem of degenerate neural text generation that occurs in language modeling \citep{holtzman2019curious}. It works by applying a penalization
term against degenerate behavior, e.g. unwanted repetitions that do not match the human training distribution, pushing down the probability of those generations.
In language modeling
this has been shown to outperform other approaches such as nucleus sampling
or beam blocking, producing state-of-the-art generations.
First experiments applying it to dialogue appear also promising \citep{li2019dontsaythat}. Finally, simply adding minimal length constraints to generations has been shown to  significantly improve human ratings \citep{roller2020recipes}.

{\em Open problems}. 
While appropriate decoding techniques have helped generative models outperform retrieval models \citep{roller2020recipes} in evaluations, this still tends to come from providing more sensible and on topic responses, rather than expressing as rich or colorful language as retrieval models, e.g. they tend to overuse common $n$-grams, and underuse rare words, and still tend to say they don't know things. Hence, to improve further, generative models should be pushed to better mimic the human distribution of training data, and to generalize that to new settings.
Besides the quality of single utterances, 
optimizing dialogue flow is a wide open problem as well.

\begin{figure}
     \centering
     \includegraphics[width=\linewidth]{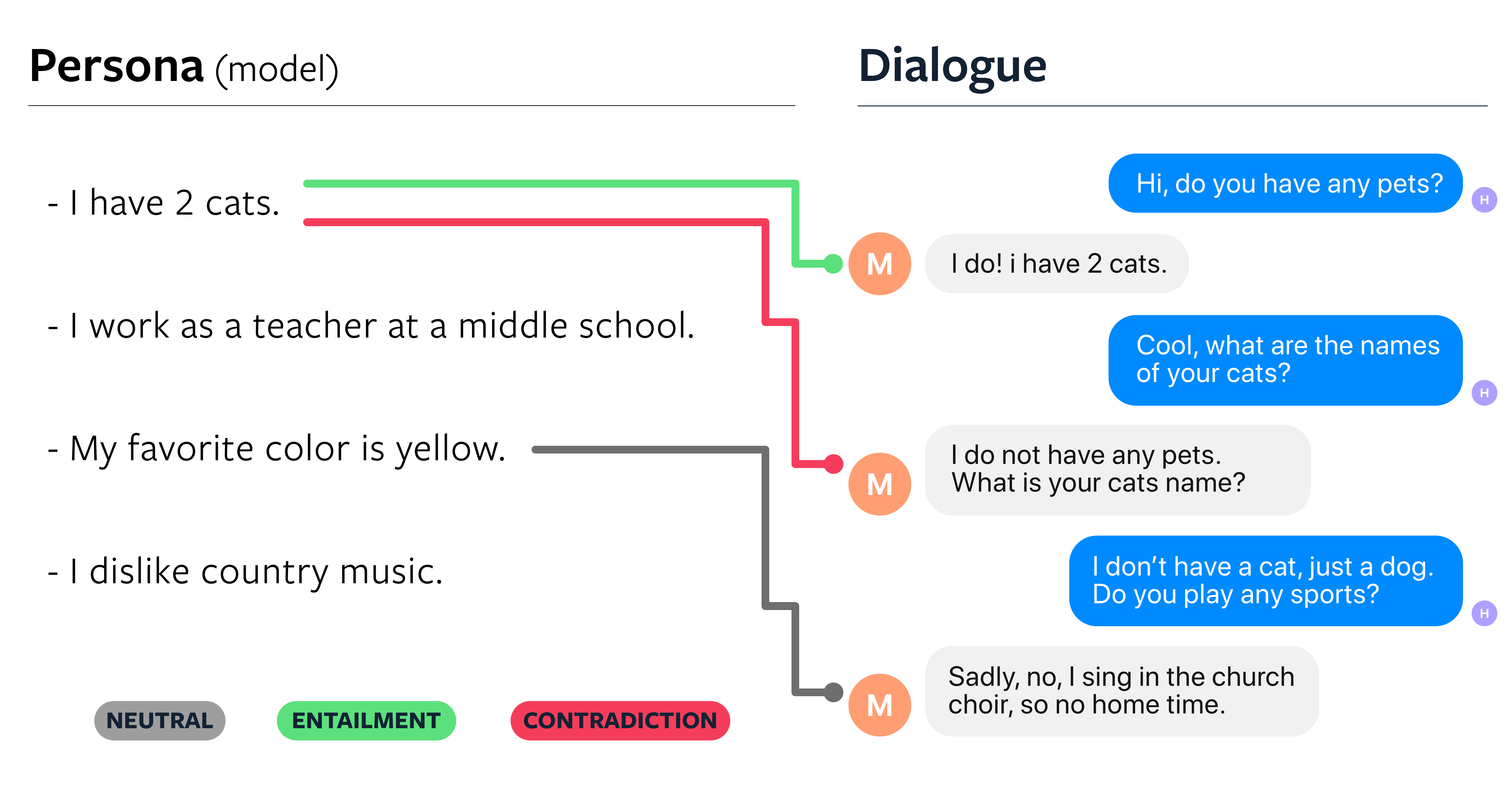}
     \caption{Dialogue natural language inference \citep{welleck2019dnli} can be used to make dialogue models more consistent.}
     \label{fig:dialogue_nli}
\end{figure}

\paragraph{Consistency}

A general problem of generative models today is that, although at first glance
the text looks human, and language modeling at the token level looks very accurate,
generating longer contexts typically exposes its flaws. While current systems are
quite good at staying on topic \citep{radford2019language}, perhaps because they still do not really 
understand what they are saying they may contradict themselves subtly or non-subtly in subsequent sentences, e.g. ``Arsenal won the premiership for the first time this year'' in one sentence and ``Arsenal have won the premiership again this year'' further on. 
While this topic is so far less studied directly in dialogue, the task of 
natural language inference (NLI) poses such understanding as a classification problem
(entails, neutral or contradicts) and progress has been made in this area 
\citep{welleck2019dnli}. Perhaps the most direct use of this research in dialogue is our work 
in developing the dialogue NLI dataset \citep{welleck2019dnli}, which directly collects such labels within
the scope of contradicting utterances in multi-turn conversations, see
Figure~\ref{fig:dialogue_nli}. We showed that training on such data and applying it as a reranker for a retrieval model decreases
the number of contradicting utterances -- across three test sets, an average of 3x fewer contradictions were observed --  while humans rated these models as more consistent and less contradictory. 
A first step in applying this same work to a generative model instead is performed in \cite{li2019dontsaythat} by applying unlikelihood training, which was described in the previous section.

{\em Open problems}. The latter work increased consistency by applying a classifier as
a post-processing operation, and within a limited domain (the Persona-Chat task, from which the Dialogue NLI dataset is derived). Future work should embed such understanding directly in the model itself so that it understands not to make these mistakes, and such understanding should generalize across many tasks. A general problem in NLI is the 
concern that classifiers are performing well by picking up on biases and shallow features rather than having fundamental understanding, and the same concerns apply here as well 
\citep{gururangan2018annotation,Poliak2018hypothesis}.

\paragraph{Memory}
Current research often does not address many aspects of memory. This is due to both our current model architectures (e.g. Transformers which condition on a small amount of input text) and our data collection procedures (e.g. crowdsourcing short conversations between strangers).
Dialogue history is typically truncated to a few turns, a ``goldfish memory'' approach, 
and even there our models do not exhibit a clear grasp of their use, e.g. the consistency issues we discussed before. 

The current approaches to using long-term knowledge are either
graph representations \citep{moon2019opendialkg} or unstructured text retrieval \citep{chen2017reading,dodge2015evaluating,dinan2018wizard}, 
which is then prepended onto the dialogue history and attended over, 
an approach advocated by the memory networks architectures \citep{weston2014memory,dinan2018wizard}. 
These approaches have been effective at
answering questions about long-term facts \citep{chen2017reading}, 
discussing topics in depth
\citep{dinan2018wizard}, and some work explores recalling 
long-term personal memories as well \citep{moon2019memory}.
For example, DrQA \citep{chen2017reading} proposed the machine reading at scale framework of retrieving
from a large unstructured knowledge base, and then performing machine reading to answer the
question, e.g. using OpenSQuAD.
Wizard of Wikipedia \citep{dinan2018wizard}, mentioned before, proposed a similar retrieval framework but for multi-turn dialogue about a topic, retrieving from Wikipedia on each turn to both answer questions, 
ask question, and respond to statements, see Figure~\ref{fig:wiz-example}.

{\em Open problems}. 
Much of this area is still open. While the latter described
 fixed knowledge base approaches are effective at utilizing 
static long-term facts, they miss two important points.
Firstly, that new memories are created all the time, i.e. there should be a {\em write}  as well 
as a {\em read} operation. We are missing both architectures, datasets  and benchmarks to train and evaluate such models.
Secondly, if knowledge is only read for a particular short-term 
goal  and never  distilled, we may limit generalization and learning.
While in machine learning some {\em read, write} memory architectures have been developed
\citep{graves2014neural,henaff2016tracking} 
they have mostly not been successful so far 
at scaling to realistic large-scale dialogue tasks.
While during training, methods like BERT \citep{devlin2019bert} do train over Wikipedia and hence 
can be shown to condense knowledge bases into their weights \citep{petroni2019language} we contend
that this is not the same as reading sentences and learning a compressed,
indexable knowledge
base (a memory) that makes generalizations from them.  For example, reading all the diverse 
information about Barack Obama and using it to build an indexable memory where this information is interrelated and some conclusions are already stored. 
Currently, any conclusions our models do make 
are simply thrown away -- e.g. the reasoning our QA systems perform every question. To build up deeper reasoning over time, presumably these need to be stored and used as building blocks -- for the model to stand on its own shoulders and think (slightly more) giant thoughts.
This also links {\em memory} to {\em continual learning}, 
which was discussed previously as an important aim.

\paragraph{Commonsense \& Reasoning}

Much of the work in conversational agents does not address reasoning directly, other
than that which is implicitly required to perform the tasks proposed. 
For task-oriented dialogue,
that ranges from understanding user utterances, to searching databases to find 
matches \citep{bordes2016learning}. Question-answering, which can be thought of as a single turn task is similar, either requiring reading comprehension \citep{rajpurkar2016squad} or retrieval as well in the more realistic case \citep{nguyen2016ms,chen2017reading}.
Although potentially any amount of reasoning is required to propose a response, many 
such tasks end up with sophisticated word overlap methods providing strong
baselines \citep{chen2016thorough}. Nevertheless, when data is in domain, systems can be built that are successful on these tasks. 

NLP researchers have thus sought to address reasoning more directly in order to 
evaluate and develop systems further. 
To do this one direction they have studied is artificial tasks  involving
controlled reasoning on toy problems in order to develop more sophisticated 
architectures \citep{weston2015towards}. This line of investigation
proved to be the first successful 
demonstration of multiple layers of
attention for reasoning with text \citep{weston2014memory,sukhbaatar2015end} which has now become part of the defacto method 
\citep{vaswani2017attention}.
Considerable resources have also been invested in developing much larger and more natural crowdsourced benchmarks such as natural language inference (NLI) tasks \citep{Bowman2015snli,Williams2017mnli,nie2019adversarial} and commonsense
reasoning tasks \citep{zellers2019hellaswag,qin2019counterfactual}.
Good progress is being made on these tasks, although questions still remain about
how much true generalization is actually occurring \citep{Glockner2018breaking,gururangan2018annotation}. Recently, an attempt to avoid
such biases has been made by collecting such tasks in rounds, where humans adversarially try to find the flaws in the models, so that they can be fixed
\citep{nie2019adversarial}.

{\em Open problems}. 
Much of the work on reasoning within the field of NLP has so far not been transferred to dialogue systems or language generation in general. 
A clear step is thus to make progress in that direction.
One intriguing possibility is to apply apply likelihood and unlikelihood training
to dialogue generation by rewarding correct reasoning and penalizing incorrect reasoning \citep{li2019dontsaythat}.

\paragraph{Multimodality and Grounding}

Language is of course often used to express concepts relating to the world we live in,
which we perceive with our eyes, ears and other senses.
Thus, grounding language to other modalities should help to learn the underlying meaning of language, and to connect to human usage. Practically, an engaging conversational agent should also
be able to discuss these other senses -- for example, the contents of an image or
a video.
Work in this area encompasses image captioning \citep{lin2014microsoft}, video captioning \citep{yu2016video}, visual QA \citep{antol2015vqa}, and more conversationally,  visual dialogue \citep{das2017visual}.
Embodied agents that use language are also being explored \citep{das2018embodied,savva2019habitat,szlam2019build,urbanek2019learning}.

\begin{figure}
     \centering
     \includegraphics[width=\linewidth]{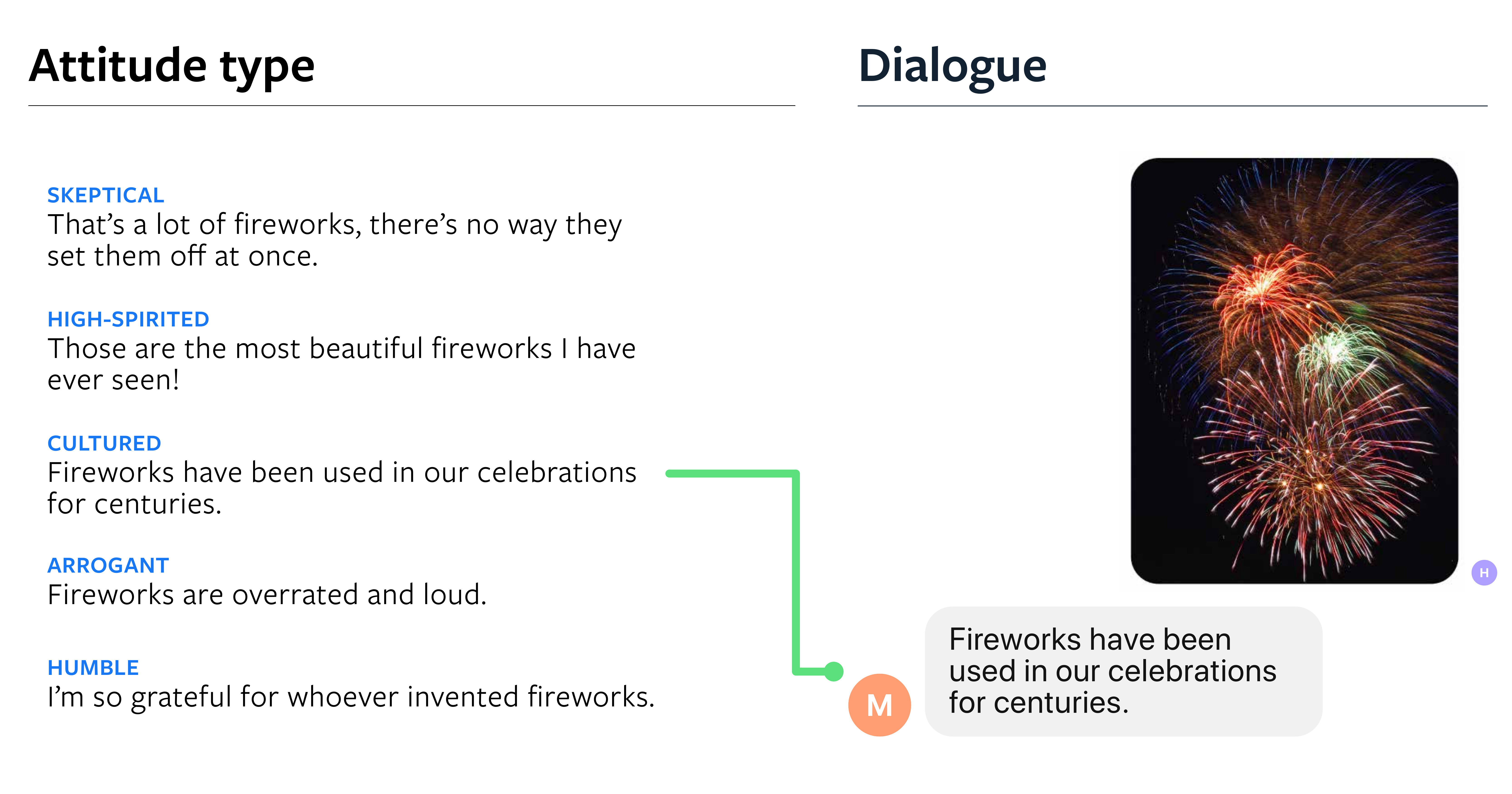}
   \caption{Conversations about Images: Image-Chat \citep{shuster2018engagingimagechat}}.
     \label{fig:image_chat}
 \end{figure}

In terms of open-domain conversation, the most relevant visual tasks are natural conversations grounded in images, such as Image-Chat \citep{shuster2018engagingimagechat} and Image Grounded Conversations \citep{mostafazadeh2017image}.
When people engage with one another and talk about what they see around them, they don’t make neutral observations — they express their points of view.  Image-Chat
is a large 187k dialogue dataset of human-human conversations about images
where the speakers incorporate given personalities, see Figure~\ref{fig:image_chat}.
In that work an architecture is developed, named TransResNet, that projects the image, personality, and caption in the same space using image (ResNet), personality, and text (Transformer) encoders. The best system is able to produce dialogue that is close to matching human performance in terms of engagement and relevance. Annotators preferred the model’s captions  on the first turn over captions written by people 49.5 percent of the time. Recent work also shows that we can combine both nonconversational multimodal data and conversational multimodal data to obtain strong performance on both \citep{ju2019allinone}.

{\em Open problems}. There is definitely less work between modalities, e.g. language and vision,  than there is of work within a single modality -- so there is much 
research to be done. We believe adding these modalities may enable 
conversational agents to be actually engaging -- as language alone does not connect so clearly with the user's direct experience. While most of this article concerns building a disembodied conversational agent, such an agent could still `see' for example by the user sending it images.  In the long-term, embodied agents either in virtual worlds or the real world via robots will be part of the picture too.

\paragraph{Personality}

Humans are strongly affected by the use of personality in language, and such language can be found to be engaging, winning over the hearts of users, independent of its other merits, such as achieving an explicit goal.

Initial attempts at training agents on dialogue data to capture personality, e.g. from OpenSubtitles movie dialogues or Twitter showed such models could express personality, but were an amalgam of all the personalities in the training set. For example asking ``what do you do for a living'' and ``what is your job?'' a single agent would answer two different professions \citep{vinyals2015neural}, related to the {\em consistency} discussion above.
In order to solve this two strategies have been tried:
trying to learn to model the personality as part of the learned weights given the speaker id from the data \citep{Li2016AModel}, or
providing training data with explicit personality information.
The latter is the subject of the Persona-Chat dataset \citep{zhang2018personalizing}, 
which consists of 1155 crowd-sourced personas, written as 5 or more sentences describing a given character, e.g.  ``I love horror movies.'', and 11k two-way conversations between randomly paired characters.
A second, larger but noisier, dataset 
where a similar type of setup has been constructed from pushshift.io Reddit has
also been built \citep{mazare2018trainingmillions,baumgartner2020pushshift}.
Persona-Chat was the subject of study of the ConvAI2 NeurIPS 2018 competition, and so is well studied by several groups \citep{dinan2019second}.

A different view of personality, rather than specific tastes and interests, is character behavior in terms of {personality traits}, 
e.g. sweet, old-fashioned or frivolous. The Image-Chat dataset, similarly to Persona-Chat, collects paired conversations with crowdworkers but this time asked to play the role of 215 such traits \citep{shuster2018engagingimagechat}.
The results show models are able to mimic such traits well with such supervised data, and that they strongly affect user engagement. For example, captions
conditioned on a personality were found to be significantly
more engaging than neutral captions, with a win rate
of 64.5\% \citep{shuster2018engagingimagechat}.

While some developers have chosen to use fixed personalities in their bots, such as Xiaoice, which has the personality of a young woman \citep{shum2018xiaoice,zhou2018xiaoice}, we believe it is
better for a bot to be able to adapt a multitude of personalities \citep{zhang2018personalizing,mazare2018trainingmillions}. Although this increases complexity, and prevents the use of well-curated copywriting, it offers a richer environment to research ideas about cognition, and enables bots with richer and more varied backgrounds. Furthermore, the ideal conversational partner is different for each user, which they may wish to choose or adapt to their desires.

{\em Open problems}. While some research progress has been made in an agent following a given specified personality, the ability to generalize from the basic description, 
e.g. if it likes one heavy metal band or one flavor of ice cream, does it like others,
has still more or less not been evaluated. Modeling these changing over time is also 
more or less unexplored, being difficult to study in the short conversation setup which is currently employed. Overall the consistency of the personality has the same issues as other types of consistency, which we discussed previously. Finally, while we can condition on
a given personality, which one of these should be matched to be engaging to a particular user, which would clearly bring gains in terms of engaging content, is also less studied.

\paragraph{Being Personal}

We make a distinction between an agent displaying personality, above, and being personal in its conversation, which we discuss here, sometimes called being {\em personalized}. 
Between humans, personal connection is important in order to build a relationship with a conversation partner.
In the beginning of a relationship, conversations often focus on simple questions about ourselves: Who are you? Where do you work? Do you have a family? Answers to these questions often drive the remainder of a conversation, as the purpose of such questions is typically to find common ground or interests. The {\em Persona-Chat} dataset directly tries to model this
\citep{zhang2018personalizing}.

{\em Open problems}.
As relationships develop, users will expect a model (or person!) to maintain a reasonable degree of continuity in just how personal the conversation is. Indeed, end-to-end chatbots are often embarrassingly forgetful, and unable to maintain even simple attributes (like a name) across multiple turns, let alone many sessions, which links to {\em memory}. Connecting dialogue research to recommendation systems research which deals with personalization also seems a clear link that should be solidified further
\citep{dodge2015evaluating,kang2019recommendation}.

\paragraph{Putting It All Together}

In order to adapt to the possibilities of different users and situations, 
all of the above aspects are important. 
Each of the individual aspects necessary for a conversational agent has unsolved, 
open problems, as we have described. 
Yet, even solving those individual problems will still leave the most important
piece -- putting them altogether into a coherent whole.

\begin{figure}[t!]
\centering
\begin{small}
\begin{tabular}{lcr}
{Human} & &  \textcolor{ourdarkblue}{BlenderBot 9.4B} \\  
\multicolumn{3}{c}{\includegraphics[width=0.44\textwidth]{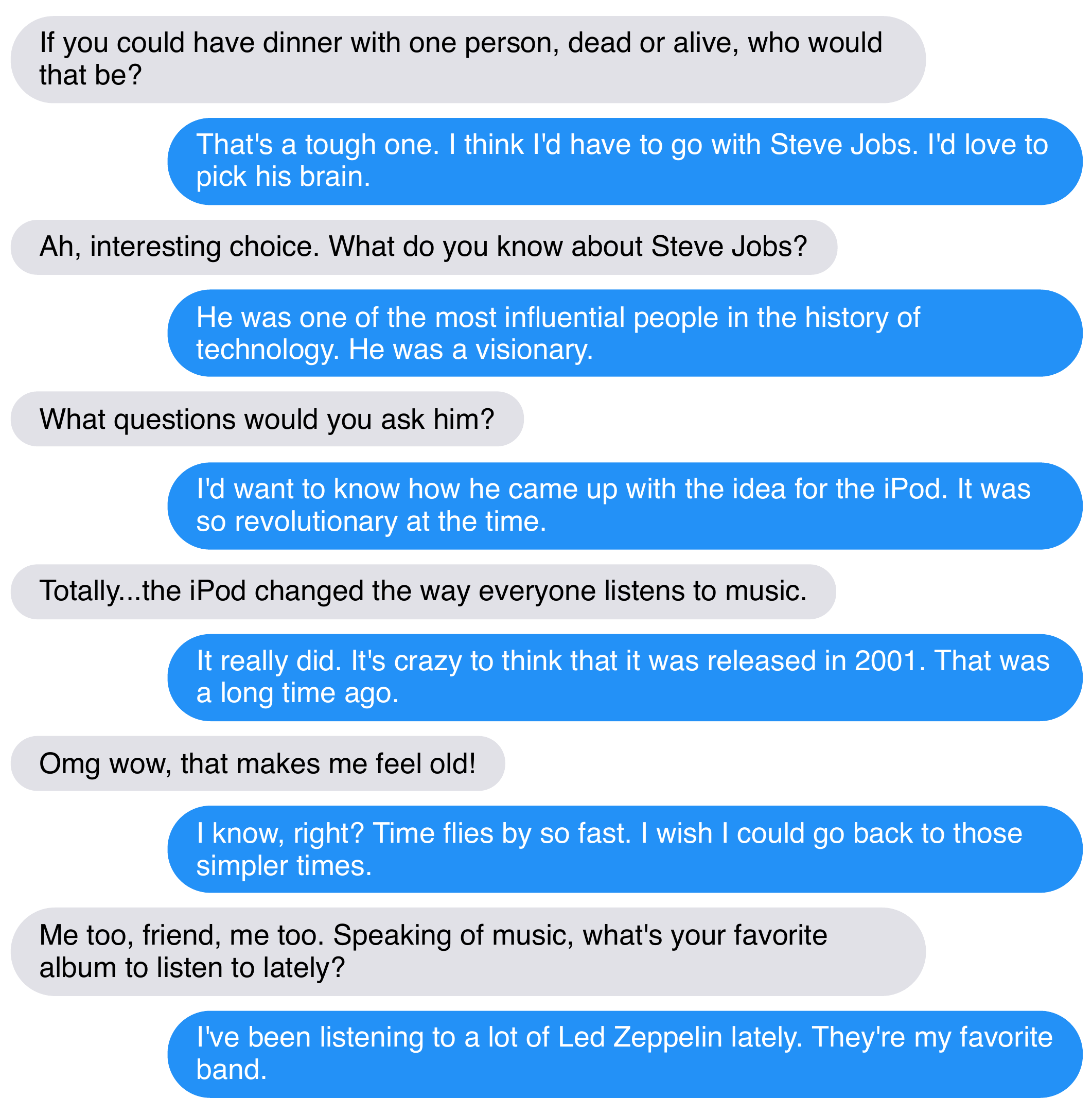}}
\end{tabular}
\end{small}
\caption{Conversation between a human and the BlenderBot model from \citep{roller2020recipes}.
\label{fig:blenderbot-example}
}
\end{figure}

To that end, a small step towards that goal has been attempted by building
a 12 dialogue task challenge, {\em dodeca}Dialogue \citep{shuster2019dialogue}. The challenge includes diverse tasks which incorporate knowledge (expertness), personality, and multimodality (images), covering some of the aspects described here.
The promise is that multi-tasking on these tasks can potentially provide a single agent capable  of all these skills, and our initial experiments indicate this
might be possible.
Taking this further, more aspects should be incorporated into such a single system,
and improved model architectures should be developed that leverage these
disparate aspects into a unified whole.
For example, \citet{smith2020bst} showed that retrieval models can seamlessly weave multiple behaviors and skills in a single conversation, including knowledgeability, personality, and empathy, by building the dataset Blended Skill Talk (BST). Such models were strongly preferred by humans over those which were only able to perform any individual skill. BlenderBot \citep{roller2020recipes} used the same strategy, first pre-training on social media data, and then fine-tuning large generative models on BST (with either 90M, 2.7B or 9.4B parameters). A cherry-picked conversation between a human and the 9.4B model is shown in Figure \ref{fig:blenderbot-example}. For more examples, including lemon-picked examples, we refer readers to that paper.

While engagingness is necessary for people to be willing to talk to a conversational agent, it is not sufficient: Tay~\citep{neff2016taybot,miller2017taybot} is an example of agent that might have been engaging, but in a way that required its removal. We now discuss points that are additional requirements for a well-behaved conversational agent.

\subsection{Well-Behaved}
An important quality for a conversational agent is to treat people the way they want to be treated. This can mean not spamming them with a deluge of unwanted messages, which is most easily accomplished by generally letting people initiate interactions. But there are also more specific caveats to take into consideration.

\paragraph{Offensive and Toxic Content}
Avoiding anything that would offend people, in terms of controversial topics, opinions, or language, while remaining engaging, is a very difficult problem. \citet{dinan2019safety} showed that it is possible to use a human-in-the-loop iterative adversarial design to improve a conversational agent along that axis through carefully designed crowdsourcing, which improved metrics on different toxic content detection tasks and made
models much more robust to adversarial attacks over three rounds of iterative refinement. Another
 finding was that the dialogue context where an
utterance appears is an important part of what makes it offensive. Other works have attempted to control for the toxicity of models by removing offensive content from the training data \citep{zhang2019dialogpt,adiwardana2020meena} or training objectives \citep{he2019negative}. It was shown in \citet{roller2020recipes} that fine-tuning on crowdworker data where workers are instructed not to use toxic language, compared to pre-training on social media data, provides less toxic models.

{\em Open problems}. Humans are very adaptable when it comes
to circumventing filters and safeguards \citep{dinan2019safety}. This is one more reason
why {\em continual learning} is important. However, there is currently
a lack of deep understanding of what makes something offensive
or objectionable to someone. Another aspect that is currently missing is how to predict people's individual preferences, both
in terms of where they draw the line between what is funny if slightly irreverent, and what is offensive, or what is
approachable, engaging language, and what is inappropriate slang.
Promising methods for controlled text generation \citep{see2019goodconversation} and text rewriting \citep{lample2018multipleattribute,smith2019zero} could be refined to provide models more
personally tailored to individual preferences,
but are still not mature enough for that application.
 Another promising route would be to train policies to avoid
 offensive or toxic utterances through reinforcement learning:
 toxic comment classifiers could be used to supply a reward signal
 and shape the conversation at a longer range than through 
 mere on-the-fly suppression. But again, it may lead to
 undesirable outcomes that models learn to only talk about
 the weather or bland topics, so reward objectives would
 have to be balanced carefully.

 \begin{figure}
     \centering
     \includegraphics[width=\linewidth]{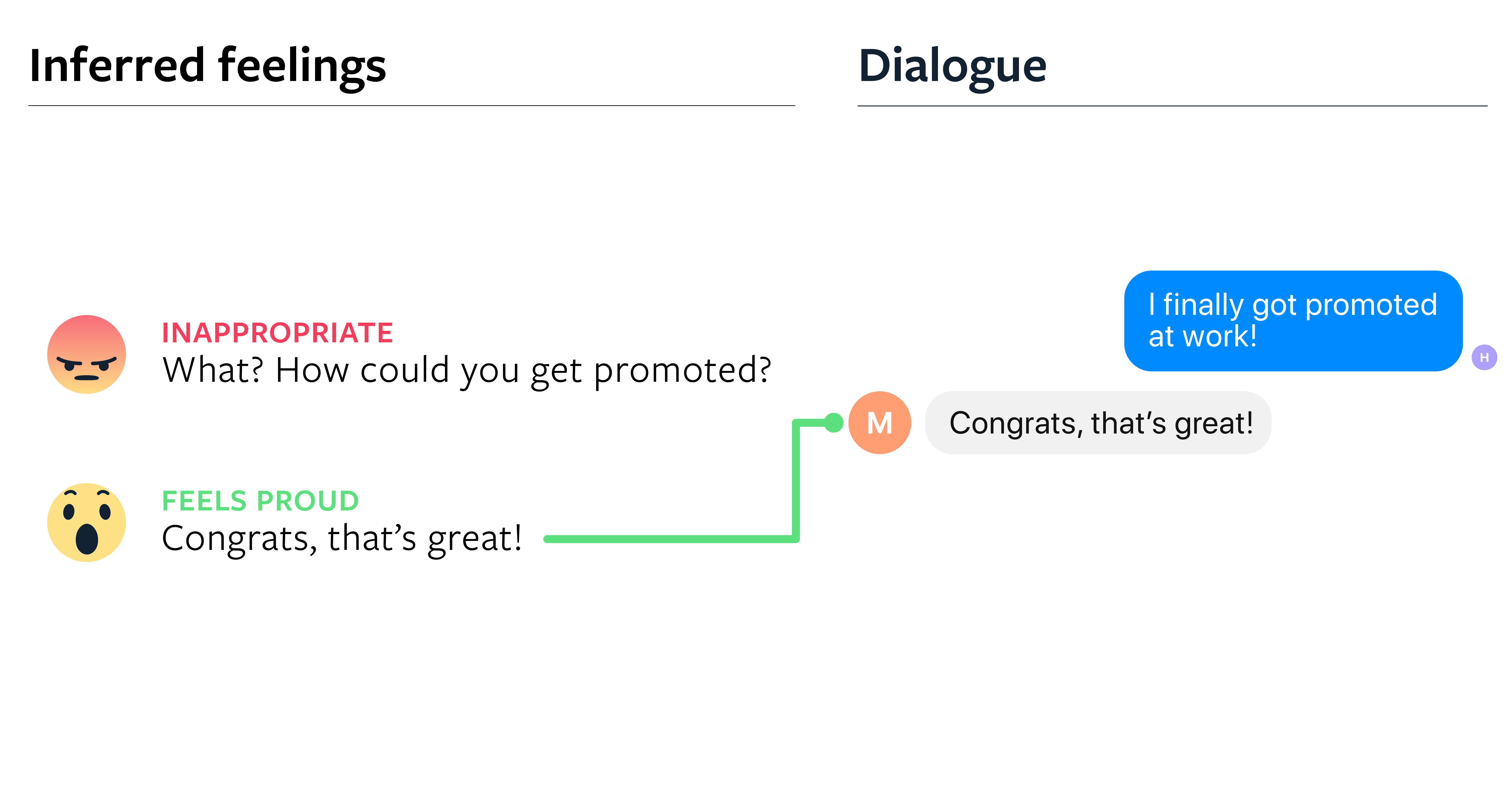}
     \caption{Empathetic dialogue \citep{rashkin2019empathetic}.}
     \label{fig:wizard}
 \end{figure}
 
\paragraph{Empathy and Compassion}
Interactions with others are received more positively and have better outcomes when they include some level of empathy \citep{wentzel1997student,levinson2000study,bickmore2001relational,kim2004effects,fraser2018spoken}, taken loosely
as
recognizing and acknowledging when the conversation partner displays
some emotion, and responding in a caring and compassionate 
manner. This is especially necessary
in open-domain conversation, which often revolves around situations that led to the experience of emotions. Since humans tend to engage with machines in a social way \citep{reeves1996media,lee2010receptionist}, it is important for conversational agents to be able to respond with empathy.
\citet{rashkin2019empathetic} proposes a benchmark and crowdsourced dataset of conversations between workers talking about situations corresponding to a balanced set of emotions, to gauge how empathetic existing models are, and shows that training models on that data yields models that are rated
as more empathetic.

{\em Open problems}. 
 While demonstrating empathy and care is
 an important objective, it is unclear how to balance it with
 other objectives such as informing or entertaining.
 While crowdworker conversations exist that contain
multiple skills \citep{smith2020bst}, these may not reflect the optimal balance we wish a final trained bot to exhibit.
It is also 
unclear whether different people
prefer different levels of empathy, and whether
this individual preference could be inferred from 
spontaneous choices of conversation topics (e.g., does the mention of a personal situation
signal a need for empathetic responding?) or otherwise signalled. If there is no universally optimal level of
empathy, a natural objective would be to be
able to control to what extent a given model
shows empathy, depending on the conversational
context and partner.

\paragraph{Privacy}
Preserving people's privacy is a central aspect of any deployed conversational agent. One approach that we have followed when deploying bot games is to frame the conversation as an artificial game where players are assigned personas \citep{zhang2018personalizing}, thus shielding their true private information through role-playing. This is a continuation
of the approach taken in multiple papers using role-played
situations \citep{rashkin2019empathetic}, assigned
traits \citep{shuster2018engagingimagechat,shuster2018imagecaption}, assigned movie preferences \citep{kang2019recommendation}, or even an entire
fantasy universe \citep{urbanek2019learning,prabhumoye2020}.

{\em Open problems}. Relying on role-playing and publicly available data creates a
potential distribution mismatch problem, where it is unclear
whether people are talking about the same things and in the same
way as if they were truly having a normal private one-on-one
conversation. This makes the creation of public benchmarks 
difficult. If improvement of an agent trained
on role-played and public data correlates well with increased
satisfaction with private conversations, then this would
be a good sign that we can keep focusing training efforts
on that regime.
Another avenue would be to explore using
privacy-preserving libraries such as CrypTen\footnote{\url{https://github.com/facebookresearch/crypten}} and decentralized approaches such as federated learning \citep{konecny2016federated} to handle learning from non-public data.
Locally personalizing a shared model (for example, on a personal mobile device) with data that would remained siloed on the personal device could be another way to deploy fine-tuned personalized models in a privacy-preserving way. 
These solutions would require drastically down-sizing current models and making them efficient and small enough that they could be loaded on device and locally updated without communicating with external servers. Benchmarks
could then rely on gauging people's satisfaction on their 
private interaction with the agent. 
Our research on more efficient models ~\citep{humeau2019polyencoder} is a step in that direction, and so are works that explore smaller footprints for Transformer-based models, e.g. through knowledge distillation~\citep{sanh2019distilbert}, adaptive spans~\citep{sukhbaatar2019adaptive}, or layer pruning~\citep{fan2019reducing}.

\section{Measuring Success}

Evaluation of Natural Language Generation remains a broadly unsolved problem, with a patchwork of solutions being used across different domains. The open-ended nature of generating sequences in a multi-turn setup naturally makes the task difficult to evaluate -- with full evaluation possessing many of the difficulties of the task itself as it requires deep understanding of the content of the conversation. In this section, we describe some of the approaches that have been used to evaluate dialogue systems, their relative advantages, and a number of open problems.

\paragraph{Human Evaluations}
Goal-oriented dialogue systems often have clear evaluation methodologies,
e.g. task completion can be measured if the correct actions are taken \citep{lemon,henderson2014second,bordes2016learning,asri2017frames,wen2016network}. Chitchat tasks, such as those discussed in this work, are more open ended, and instead feature conversations without a precise goal that can be automatically evaluated. Furthermore, automatic metrics (discussed below), have not been shown to have a clear correlation with human evaluations \citep{liu2016not,lowe2017towards}. This means the current standard for all dialogue research involves human trials.

However, there are multiple ways one may choose to evaluate the effectiveness of the system, and human judgements are often difficult to measure. Today, the two most common evaluation forms for dialogue include single-turn pairwise evaluation, and multi-turn Likert evaluation.

In single-turn pairwise evaluation \citep{vinyals2015neural,Li2016AModel}, a human evaluator is typically presented with a full conversational context, and  shown two possible responses, and asked to pick which model they feel is better. This test affords the benefits and simplicity of an A/B test, but fails to take into account any multi-turn aspects of a conversation. For example, a model which repeats itself across multiple turns will not be identified by such a system, a behavior known to be highly disliked by human evaluators \citep{see2019goodconversation}. It furthermore removes any noise produced across multiple turns, wherein a system would be required to ingest its \emph{own} responses in the conversation history, rather than some produced by a human \citep{li2019acute}.

Another common evaluation framework is multi-turn Likert evaluation \citep{ram2017alexaprize,venkatesh2018evaluating,zhang2018personalizing,rashkin2019empathetic,see2019goodconversation,dinan2019second,dinan2018wizard}, in which a human evaluator is asked to discuss with an agent for several minutes, and then evaluate performance on a Likert (1--5) scale. Such evaluations easily capture a model's ability to carry on longer conversations, and handling of out-of-distribution situations, and therefore may be preferred over single-turn pairwise evaluations. However, multi-turn Likert is not without its own difficulties: it is considerably more labor intensive than A/B tests, as it requires longer and higher-cognitive involvement from the annotators, and it relies on absolute identification rather than relative discrimination, even though absolute identification is not reliable in humans \citep{stewart2005absolute}. Likert evaluations are often not strong enough to find statistically significant differences between some models, making it difficult to measure incremental improvements \citep{kulikov2018importance}. To make matters worse, it is usually necessary that one must also re-evaluate the baselines at the same time as one's novel model, as the distribution of human annotators can easily shift over time, causing measurement errors \citep{see2019goodconversation}. Another common difficulty is related to sequential effects (e.g., reviewed in \citet{stewart2005absolute}),
where the first system an annotator evaluates can heavily influence their future ratings, causing difficulties in using an absolute scale.

\begin{figure} 
    \centering
    \includegraphics[width=\linewidth]{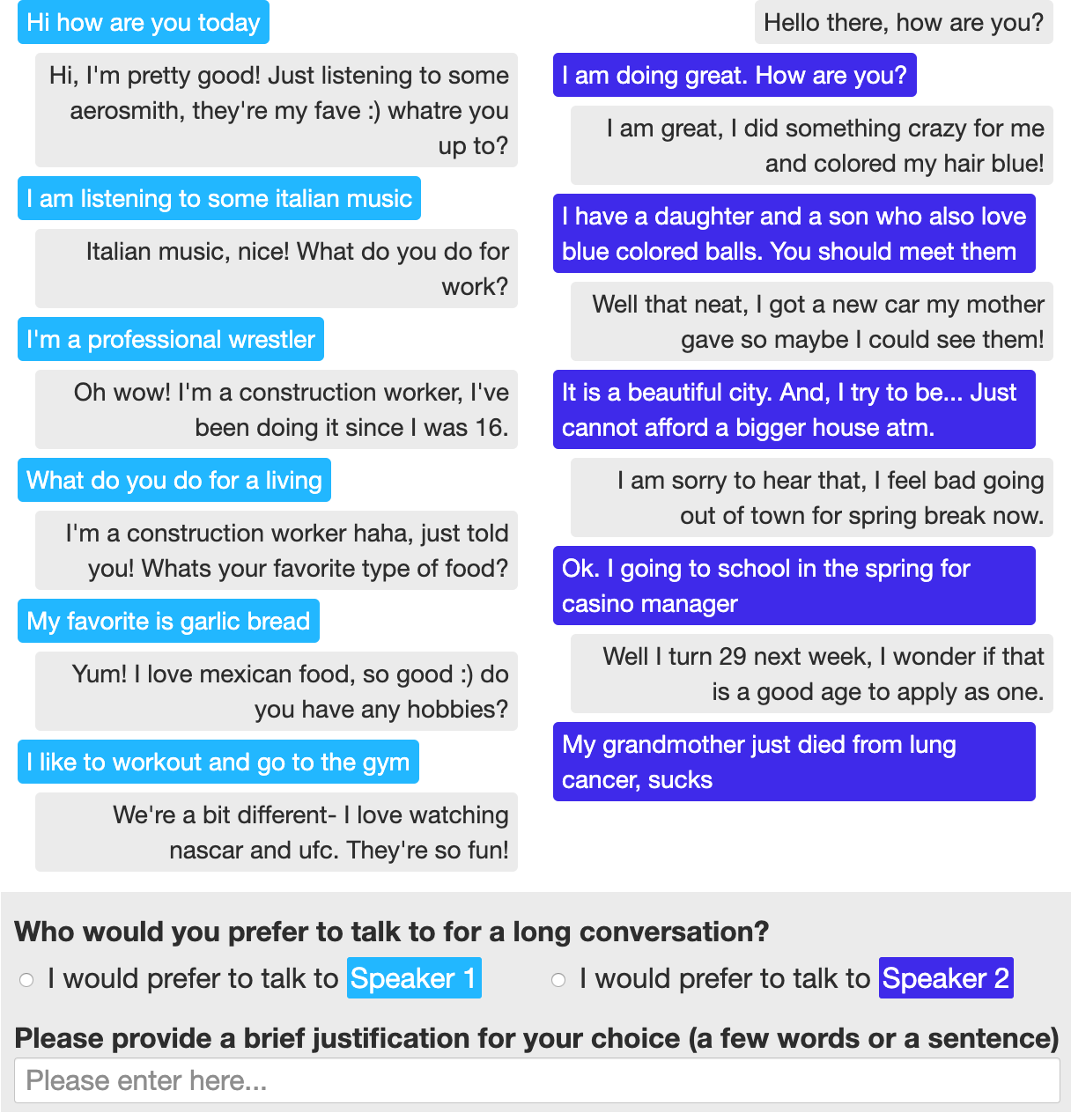}
    \caption{ACUTE-Eval has human annotators directly compare multi-turn conversations with different systems.}
    \label{fig:acute}
\end{figure}

Some groups have proposed hybrid approaches between single-turn pairwise evaluation and multi-turn Likert scoring. For example, \citet{novikova2018rankme} propose a method that combines continuous scales and relative assessments, but in single-turn, rather than multi-turn evaluation; and \citet{adiwardana2020meena} propose binary good/bad annotations of individual utterances in a multi-turn setting. \citet{li2019acute} recently proposed ACUTE-Eval, in which evaluators are asked to complete pairwise evaluations of complete dialogues. An example of ACUTE is shown in Figure~\ref{fig:acute}. This setup affords a number of advantages over both single-turn pairwise, and multi-turn Likert evaluations. The explicit use of comparisons remedies many of the issues of sequential effects, while still providing the ability to expose issues that are present only in multi-turn evaluations.

Furthermore, the pairwise setup facilitates replication and efficient reuse of data: conversations collected in previous trials and by other systems can be directly compared with a new system, without having to recollect additional data. This can significantly reduce the resources needed by a new evaluation, and ensure that multiple papers are comparing to prior work consistently.

As a trade-off, ACUTE-Eval does require that one performs \emph{two} stages of evaluation: one where humans conduct conversation with a model, and another where third-persons indicate pairwise preferences. If one has many systems to compare, this may actually increase resource requirements, since one must pay the full price of multi-turn collection, and another of pairwise evaluations. Fortunately, we can reuse the same dialogue in multiple pairwise comparisons, reducing the number of conversations required to detect statistical significance, alleviating some of the issue. When comparing to multiple existing systems, the benefit of being able to re-use old collections outweighs the resource requirements of the new collections, mitigating these effects \citep{li2019acute}.

However, as an alternative, we find that ACUTE-Eval can also work in ``self-chat'' mode, where models are used for \emph{both} sides of a conversation, instead of human-model chat. This eliminates the requirement of the initial human collection, and conversations may be generated without human involvement, dramatically reducing the resource requirements of evaluation. We found in our experiments that results from self-chat experiments highly correlated with those of human-chat experiments, for most, but not all systems \citep{li2019acute}. This mirrors other successes in using self-play, self-chat, and simulated users to evaluate dialogue systems \citep{fazelzar2017learning,shah2018bootstrapping,shah2018building,wei2018a,ghandeharioun2019approximating}.

\paragraph{Automatic metrics}
Evaluation of chitchat tasks with automatic metrics is difficult precisely because of their open-ended nature. For example, the answer to the question ``What are you doing tonight?'' has many possible answers, each with little word overlap. This means standard metrics based on word-overlap with reference responses, as frequently used in question-answering \citep{rajpurkar2016squad} or machine translation \citep{papineni2002bleu}, do not work well, and have poor correlation with human judgments \citep{liu2016not,novikova2017we,lowe2017towards}.
Nevertheless, a number of studies do report automatic metrics, sometimes without human studies \citep{lowe2015ubuntu,serban2016building,parthasarathi2018extending}. Some commonly used word-overlap metrics include F1 \citep{rajpurkar2016squad}, BLEU \citep{papineni2002bleu,li2017dailydialog}, ROUGE \citep{lin2004rouge,fan2019eli5}, CIDEr \citep{vedantam2016cider,zhou2018xiaoice}, and METEOR \citep{banerjee2005meteor,zhou2018xiaoice}. Each covers slightly different aspects, and may be more appropriate in specific situations, but none is known to be a perfect evaluation of conversational models.

More specialized metrics may be used for specific subtypes of conversational AI systems. For example, ranking models are often evaluated using Recall @ K or Top-1 Accuracy \citep{zhang2018personalizing,dinan2018wizard, dinan2019second,humeau2019polyencoder}. These can be used as rough proxies for improvements, but the open nature of dialogue means that there may be many valid answers in a given candidate list. Such metrics are also unable to capture how well a model will \emph{generalize} to new situations.

Similarly, generative models typically report perplexity of a held-out test set (e.g. \citet{li2017dailydialog,dinan2019second,shuster2019dialogue,fan2019eli5,adiwardana2020meena,roller2020recipes}), and recent work has even found perplexity correlates strongly with human evaluations within the same model class \citep{adiwardana2020meena}. While perplexity does give a good estimate of the probability that a generative model would produce the gold label, such results may be actually quite rare under beam search \citep{fan2018controllable,holtzman2019curious,welleck2019neuraltext}, and not representative of an actual generation of a model under beam search or sampling. Perplexity also depends on the dictionary, and not all models will necessarily have entirely comparable perplexities, especially when unknown words are present in validation or test labels, making it difficult to compare systems across multiple time horizons \citep{dinan2019second}. Modern model using BPE dictionaries further complicate complications of comparing perplexities across multiple systems \citep{sennrich2016bpe}. Specialized systems, which focus on improving specific behaviors of generative models, might instead focus on specialized metrics that are not indicative of overall generation quality, but instead on specialized behavior like repetition \citep{see2019goodconversation,welleck2019neuraltext,li2019dontsaythat} or vocabulary usage \citep{see2019goodconversation,holtzman2019curious,li2019dontsaythat}. Altering the behavior of the generation method can dramatically influence human evaluations, while maintaining identical or near-identical perplexity \citep{see2019goodconversation,welleck2019neuraltext,welleck2019dnli,adiwardana2020meena,roller2020recipes}.

Noting the inadequacy of each of these automatic metrics, a number of researchers have proposed \emph{learning} metrics for dialogue evaluation \citep{lowe2017towards,ghandeharioun2019approximating}. Typically, this is done via a regression from a number of automatically-extracted features (e.g. sentiment and semantic similarity) to human evaluations. In particular, \citet{ghandeharioun2019approximating} perform such correlations using features extracted via self-play of models.
Such systems provide a promise of improved speed of research and development of dialogue agents, but so far have not been met with wide adoption. A common point of criticism is that there can be little effective difference between a learned metric and one that is used as an actual model for performing utterance selection. Put another way, one can easily maximize a metric by employing methods like ranking all utterances according to a learned metric, or using Monte Carlo Tree Search \citep{kumagai2016mtcs} during generation to  na\"ively optimize the automatic metric. In this manner, the problem of learning an automatic metric is difficult to disentangle from the rest of dialogue research.

\emph{Open problems}. Selection of an automatic metric for dialogue research, or natural language generation in general, remains a widely open problem that attracts many researchers. Despite concerns, we remain optimistic about methods which approach the problem via learning. Future work may additionally consider holistic evaluations, which require the full conversation to complete before being able to make an individual prediction. This may help mitigate concerns around using the metric to drive dialogue selection. Similarly, adversarial networks may provide a potential avenue for improving automatic selection via continually improved detection of compute-generated responses.
In the short term, shared tasks may offer the best avenue to finding automatic metrics which correlate with human judgements \citep{ram2017alexaprize,dinan2019second,yoshino2019dialog}, but also rely on a diversity of submitted systems in order to consider such evaluations. If all participants use similar models with similar pretraining using similar corpora, then we should not expect clear distinctions to be made apparent in shared tasks.

\paragraph{Behavioral Metrics}
Yet more alternatives are available as models are deployed to real users, especially behavioral metrics. For example, in the Alexa Prize, models were evaluated by how many turns were completed before a conversation was abandoned \citep{ram2017alexaprize}. Others might be evaluated by the retention rate of users (e.g. how many users \emph{choose} to have a second or third conversation with a model). Such behavioral metrics can be powerful implicit indicators of preferences of users, but have a large number of issues. For example, models which frequently ask for clarification will naturally result in more turns of conversation, but naturally frustrate users; systems which initiate a conversation will have higher retention, but may not be appreciated by users. A careful and thoughtful balance of allowed behaviors must be employed, and researchers should feel discouraged from using ``engagement hacks.''

\emph{Open problems}. There is significant question as to what are the correct implicit behavioral metrics to collect, and what few methods exist now depend heavily on the medium and design of the system. As more models are deployed to the wild, we encourage researchers to share their successes and best practices so that the community may come to a consensus.

\paragraph{Discussion}
It is likely that all of the above (human evaluations, automatic metrics, and behavioral metris) and more, will need to be measured with some granularity, in order to understand trade-offs of different behaviors and attributes. In the short term, deployed models should likely focus on retention, in order to ensure a steady stream of users to afford experimentation and iteration.

\section{Discussion}

In this section, we strive to enumerate our core research and ethical values. We discuss how we prioritize trade-offs in our decisions, as well as lessons internalized from our experiences with different steps in the development process. We end on reflections of trends in the community, and calls for action within the community.

\subsection{Values and Guiding Principles}

One primary principle behind our work is \emph{openness}. We strive, whenever possible, that the findings of our research should be shared freely via publications whenever it provides benefit. Furthermore, the items necessary to reproduction of our results should additionally be made public when possible. This includes pretrained models, but also code and data necessary to reproduce these results. We believe that siloed research inhibits progress of the field, and point to the recent radical improvements in the NLP community stemming from the openness of the publication of Transformers \citep{vaswani2017attention} and explosion of following open models \citep{devlin2019bert,lample2019cross,dai2019transformerxl,yang2019xlnet,liu2019roberta}. With the trend of pretraining coming to dominate the field, open datasets are more important than ever. Our own data, models, and code will are made public via ParlAI\footnote{\url{https://parl.ai}} \citep{miller2017parlai}, our unified platform for dialogue research. Our current best approach, BlenderBot \citep{roller2020recipes} is available there.

Indeed, our unified platform additionally propels our second value: \emph{swiftness}. In the past few years, the NLP community has been radically changed via massive improvements to the availability of compute and data. Reacting and improving upon the most state-of-the-art work will be important to the success of open-domain conversational agents. We have found ParlAI to be important to remaining swift during these times of rapid development. By providing a unified platform for collection of data, implementation of models, evaluation of agents, and deployment of agents, we are able to significantly reduce development time. For example, our recent development of Polyencoders \citep{humeau2019polyencoder} required no modification to be evaluated using our new evaluation framework \citep{li2019acute}.

We also prioritize \emph{privacy} as a value in our research. Online chats are often where our most intimate and sensitive discussion happens, and one may imagine that users may be even more uninhibited in their interactions with bot. As such, we must act responsibly with respect to data releases. This means that all users must be provided informed consent around how their conversations will be used. Furthermore, we should only release anonymized versions of data, and make every effort to ensure that sensitive data is not included in public releases. Indeed, we must always value privacy over openness and swiftness, whenever our values are in direct conflict with one another. However, we believe that we can have all three values at once: for example, games with role-playing aspects, like Beat-the-Bot, have mitigated the likelihood of sensitive information being included in a conversation, and enable us to open-source the data. In future deployments, we will also add a private mode to appropriate selections, which disables logging and data collection. We also hope that federated learning \citep{konecny2016federated} and other privacy-oriented machine learning techniques will enable future variants to perform learning in a privacy-first manner.

\subsection{Our Experiences}
We have internalized a number of lessons from experiences with training and releasing models.

\paragraph{Pretraining} First, we have found that pretraining is important to performance for nearly every variant of chitchat we have experimented with \citep{wolf2019transfer, dinan2019second}, including both in retrieval \citep{humeau2019polyencoder} and generative methods \citep{shuster2019dialogue,zhang2019dialogpt,adiwardana2020meena,roller2020recipes}. Furthermore, we have consistently found that domain-specific pretraining is important to high performance: that is, using dialogue-like pretraining significantly outperforms generic pretraining on resources like Wikipedia and BooksCorpus \citep{dinan2018wizard,humeau2019polyencoder,dinan2019safety,dinan2019second}. This further underscores the importance that our models should be openly available, in order to ensure researchers with fewer computational resources are still able to conduct high-quality research. Such efforts are important to ensuring pretraining acts as a rising tide, rather than protectionism for the groups with the most resources.

\paragraph{Efficiency} Even groups with large computational resources will find that models must be computationally accessible in order to be deployed on a wide scale. Deployed models need to run on commodity hardware, without access to GPUs or TPUs. Indeed, this was the core motivation behind the development of Polyencoders \citep{humeau2019polyencoder}. As a rule of thumb, a researcher should be able to communicate with her model in real-time on a mid-tier laptop, with zero additional development effort. This restriction ensures that we are developing models that are able to be deployed easily. Furthermore, since automatic metrics are untrustworthy in dialogue, it also ensures that a researcher can manually test her model, understanding its power and limitations. Although the recent trend in NLP is to train larger models requiring GPUs for inference \cite{devlin2019bert,liu2019roberta,radford2019language,zhang2019dialogpt,adiwardana2020meena}, methods for more efficient inference and smarter algorithms provide ways to retain performance while keeping high performance \citep{sanh2019distilbert,fan2019reducing,humeau2019polyencoder}.

\paragraph{Best practices} We have also adopted a number of software engineering best practices around our development process. In particular, we have found automatic testing and continuous integration to be invaluable to our developments. We regularly verify our released models perform as expected, allowing us to readily identify and remedy backwards-compatibility issues. High quality code reviews, even during early model prototypes, have helped us identify bugs and misunderstandings in our models early and sped up development. Universal usage of a shared platform ParlAI
\citep{miller2017parlai} has ensured that we maintain a consistent level of quality across multiple projects, and that we can minimize efforts around reproduction and guarantee performance and longevity of models. Many of these benefits are obvious and well-known to software engineers, but are easily forgotten and ignored in research.

\paragraph{Deployment} In contrast to development of models, \emph{deployment} has presented us with a very different set of lessons. There are a number of major engineering complications involved whenever games require the involvement of {\em two} or more humans, as in Beat-the-Bot; these run antithetical to usual scaling recommendations like sharding. Slowness in pairing can significantly frustrate users, and cause them to abandon the game, further exacerbating the issue. Furthermore, users want the game to react instantly, but wish to take their time in responding. The result is that users may be more satisfied with games and interactions that do not require another human to be present.

We have also found that deploying models has resulted in a consistent and steady stream of adversarial users (``trolls''). This is highly unsuprising, as shown by the legacy of Microsoft's Taybot \citep{neff2016taybot,miller2017taybot}. Interestingly, we found that adversarial users had a tendency to assume that they were training the bot online, similar to Microsoft Tay, and believed the bot would learn to mimic their suggested responses quickly. These users focused heavily on suggesting highly offensive responses, especially to innocent and common questions.
Other adversaries focused on asking sensitive questions in hopes the bot would produce an offensive response, with the intention of publicizing the bot's failures. Both of these experiences emphasize the importance of the safety of responses, especially around the need for safety-in-context. They also demonstrate the significant risks in online learning, and why it must be deployed with extreme caution and safeguards.

\subsection{Recommendations to the Community}

Based on our experiences and view of open problems, there are a number of recommendations we suggest to the community, in hopes of encouraging development while continuing the pursuit of open and accessible science.

\paragraph{Shared Tasks} We urge the community to rally behind a definitive set of tasks and measurements for conducting engaging chitchat research. The current state of the fractured community makes it difficult to compare works, despite having similar domain and purpose. Recent competitions, such as the ConvAI2 challenge \citep{dinan2019second} and the DSTC7 challenge \citep{yoshino2019dialog}, stand as excellent models for such endeavors. As we progress, these challenges must incorporate more and more difficult challenges encompassing all of the behaviors that are necessary for the ultimate bot. We note that our recently developed DodecaDialogue suite \citep{shuster2019dialogue} offers an evaluation framework that encompasses many of the research challenges discussed in this document, and encourage the rest of the community to employ it.

Such standardized tasks should also include real, interactive learning systems, such as those developed in the Alexa Prize competition \citep{ram2017alexaprize}. We hope that future iterations will also provide for more liberal open data and participation. We believe these are an excellent way for research to progress, and encourage these continue and expand.  Naturally, this is complicated by the number of open research problems, such as what is the correct way to do automatic evaluation of systems.

\paragraph{Software} As we standardize on data, groups may see additional benefit from standardizing on software stacks as well. Our ParlAI \citep{miller2017parlai}\footnote{Available at \url{https://parl.ai}} framework attempts
to integrate standardized data, released models and software
as a one-stop solution and its wide adoption by the community
has created an ecosystem of improved support, feature
development, and engineering improvements. Our openness
to collaborations, contributions, and requests from groups
outside our own organization is a reflection of our
core belief that sharing research tools is the most
productive way to make fast advances as a field.

A number of design principles went into ParlAI, which we believe have paid repeated dividends and enabled faster research for ourselves. In particular, ParlAI has focused on unifying the format of all {\em tasks} into simple input-output text pairs have helped us to treat ensure our systems are highly reusable, and that different tasks may be easily combined to produce models which exhibit joint behaviors. We also attempt to avoid overly-specialized architectures as often as possible, ensuring our models are also useful across a wide variety of tasks. Furthermore, we observe that today's best models are tomorrow's baselines, and that the centralized repository of reproducible and pretrained models and implementations significantly lowers the overhead needed to perform comparisons to prior work.

Perhaps more easily overlooked is the success we have experienced by enforcing \emph{all} parts of the system act via a unified API of \emph{Agents}. This means that datasets (teachers), models, crowdworkers, and actual users are all exposed through a unified means. As a result, we may move effortlessly and raplidly progress between Wizard of Oz data collection, model training, human evaluation, and model deployment.

Other efforts in the software space include RASA \citep{bocklisch2017rasa}, PolyAI \citep{henderson2019repository}, Uber's Plato \citep{papangelis2019collaborative}, Microsoft's IceCaps \citep{shiv2019microsoft}, and Huggingface Transformers
library \citep{Wolf2019HuggingFacesTS}, not focused on
dialogue per se, but used as a base in much dialogue work.

\section{Conclusion}

Our research has shown that it is possible to train models to improve on some of the most common weaknesses of chatbots today. Over time, we’ll work toward bringing these subtasks together into one unified intelligent agent by narrowing and eventually closing the gap with human performance. In the future, intelligent chatbots will be capable of open-domain dialogue in a way that’s personable, consistent, empathetic, and engaging.

As part of our contribution to the broader research community, we’re sharing our new models, training code, and data sets within ParlAI \citep{miller2017parlai}, our open source dialogue research platform. We hope that this platform will continue to foster research advances across the research community and contribute to pushing dialogue research forward, addressing many of the open problems we have described here.

\bibliography{main}
\bibliographystyle{aaai}

\end{document}